\documentclass{article}

\usepackage[numbers]{natbib}

\usepackage[preprint]{neurips_2020}

\usepackage[utf8]{inputenc} 
\usepackage[T1]{fontenc}    
\usepackage{hyperref}       
\usepackage{url}            
\usepackage{booktabs}       
\usepackage{amsfonts}       
\usepackage{nicefrac}       
\usepackage{microtype}      
\usepackage{amsmath}
\usepackage{subcaption}
\usepackage{tikz}
\usetikzlibrary{bayesnet}
\usepackage{tabularx}
\usepackage{float}
\usepackage{caption} 

\newcommand{\bx}{\textbf{x}}
\newcommand{\bX}{\textbf{X}}
\newcommand{\bu}{\textbf{u}}
\newcommand{\bU}{\textbf{U}}
\newcommand{\bh}{\textbf{h}}
\newcommand{\bz}{\textbf{z}}
\newcommand{\bb}{\textbf{b}}

\newcommand{\bmu}{\boldsymbol{\mu}}
\newcommand{\bsigma}{\boldsymbol{\sigma}}
\newcommand{\bSigma}{\boldsymbol{\Sigma}}
\newcommand{\diag}{\mathop{\mathrm{diag}}}

\usepackage[most]{tcolorbox}

\newtcolorbox{highlighted}{colback=yellow,coltext=black,breakable}
\usepackage{framed}
\colorlet{shadecolor}{yellow!100}
\usepackage{algorithm}
\usepackage{algpseudocode}

\title{Recurrent Flow Networks: A Recurrent Latent Variable Model for Density Modelling of Urban Mobility}

\author{%
  Daniele Gammelli \\
  Technical University of Denmark\\
  \texttt{daga@dtu.dk} \\
   \And
  Filipe Rodrigues \\
  Technical University of Denmark\\
  \texttt{rodr@dtu.dk} \\
}

\begin{document}

\maketitle

\setcounter{page}{20}

\begin{abstract}
Mobility-on-demand (MoD) systems represent a rapidly developing mode of transportation wherein travel requests are dynamically handled by a coordinated fleet of vehicles.
Crucially, the efficiency of an MoD system highly depends on how well supply and demand distributions are aligned in spatio-temporal space (i.e., to satisfy user demand, cars have to be available in the correct place and at the desired time).
To do so, we argue that predictive models should aim to explicitly disentangle between \emph{temporal} and \emph{spatial} variability in the evolution of urban mobility demand.
However, current approaches typically ignore this distinction by either treating both sources of variability jointly, or completely ignoring their presence in the first place.
In this paper, we propose \emph{recurrent flow networks}\footnote{Code available at \url{https://github.com/DanieleGammelli/recurrent-flow-nets}} (RFN), where we explore the inclusion of (i) latent random variables in the hidden state of recurrent neural networks to model temporal variability, and (ii) normalizing flows to model the spatial distribution of mobility demand.
We demonstrate how predictive models explicitly disentangling between spatial and temporal variability exhibit several desirable properties, and empirically show how this enables the generation of distributions matching potentially complex urban topologies.
\end{abstract}

\section{Introduction}
\label{sec:introduction}
With the growing prevalence of smart mobile phones in our daily lives, companies such as Uber, Lyft, and DiDi have been pioneering Mobility-on-Demand (MoD) and online ride-hailing platforms as a solution capable of providing a more efficient and personalized transportation service.
Notably, an efficient MoD system could allow for reduced idle times and higher fulfillment rates, thus offering a better user experience for both driver and passenger groups. 
The efficiency of an MoD system highly depends on the ability to model and accurately forecast the need for transportation, such to enable service providers to take operational decisions in strong accordance with user needs and preferences.
However, the complexity of the geospatial distributions characterizing MoD demand requires flexible models that can capture rich, time-dependent 2d patterns and adapt to complex urban geographies (e.g. presence of rivers, irregular landforms, etc.).
Specifically, given a sequence of GPS traces describing historical demand patterns, obtained through e.g., logging of mobile-app accesses, service providers are challenged with the task of forecasting the 2d spatial distribution, i.e. in longitude-latitude space, of future mobility demand.

\begin{figure}[t]
\normalsize
\centering
\includegraphics[width=\columnwidth]{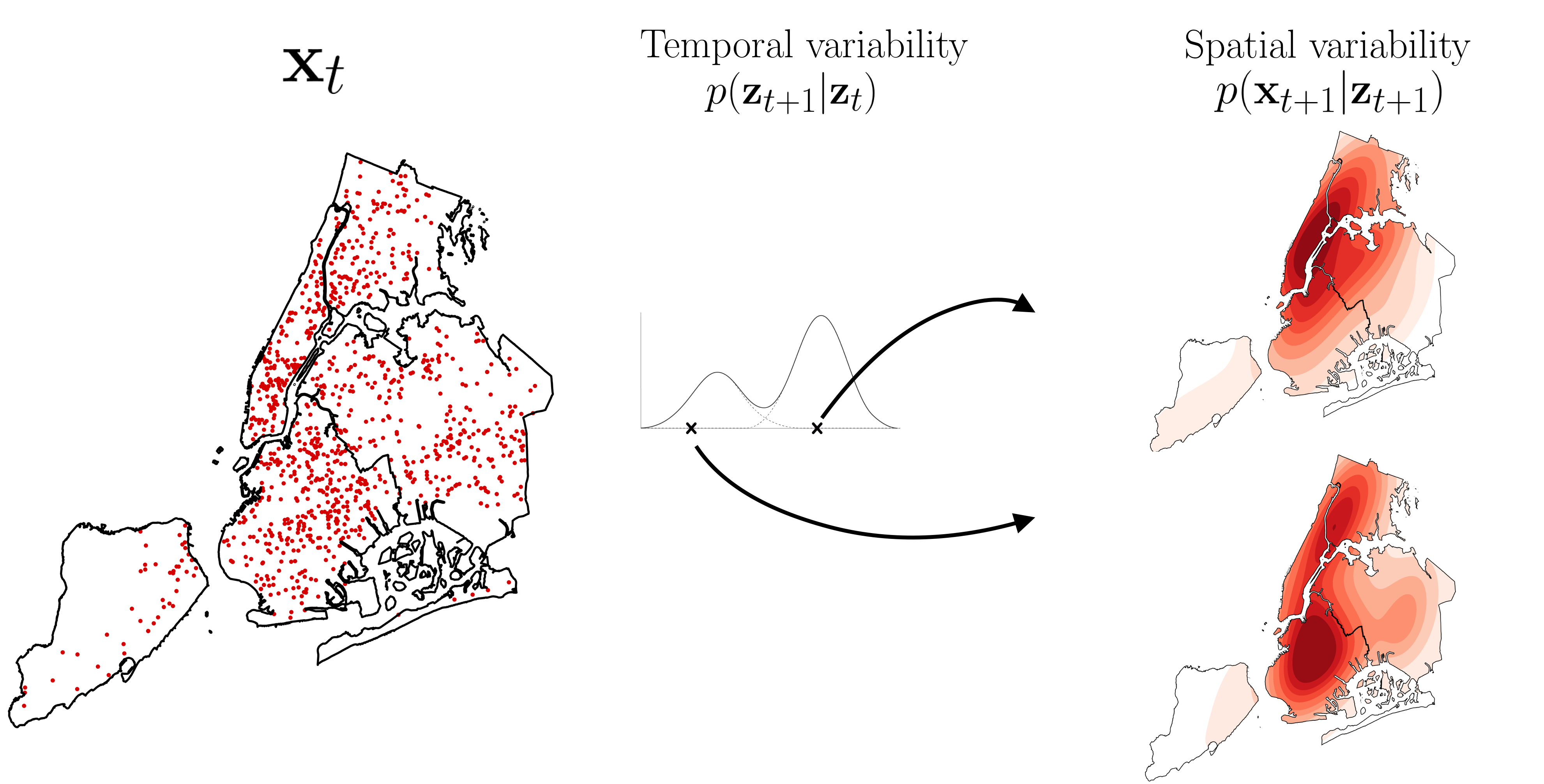}
\caption{This paper proposes a framework to estimate flexible spatio-temporal distributions of urban mobility. Given the current positioning of mobility demand ($\bx_t$), there are multiple possible future scenarios for its evolution. We encode this \emph{temporal variability} through latent random variables ($\bz_t$), in order to capture potential multi-modalities of the unobserved demand generating process (e.g. different scenarios based on special events, extreme weather conditions, etc). For every sample from the learned latent process $p(\bz_{t+1} | \bz_{t})$, we encode the \emph{spatial variability} through normalizing flows and show how this additional flexibility enables statistical models to better estimate distributions over complex urban topologies (e.g. rivers, parks, etc.), crucial for downstream decision-making tasks.}
\label{fig:contribution} 
\end{figure}

\begin{figure}[t]
\normalsize
\centering
\includegraphics[width=\columnwidth]{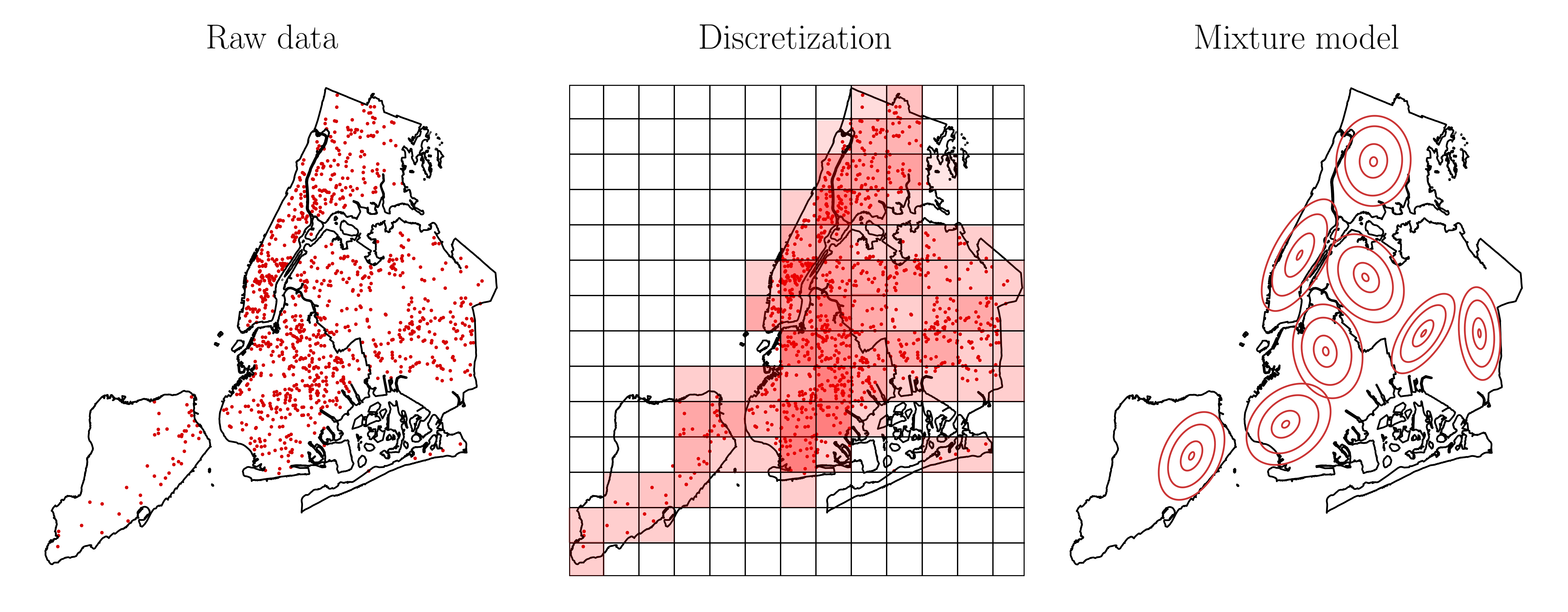}
\caption{Given raw GPS traces (i.e., longitude-latitude), we argue that the design of deep learning architectures for the task of urban mobility modeling should focus on avoiding both spatial discretizations (center) and mixture-based output distributions (right).
Opposed to e.g. pixel values in an image, GPS traces are naturally defined as continuous observations, and a discrete grid might be unable to represent real urban geographies. 
Following a similar reasoning, despite being defined on a continuous support, mixture models might also exhibit structural limitations when dealing with geographical data (e.g., choice of the number of mixture components).}
\label{fig:contribution2} 
\end{figure}

\smallskip \noindent In this work, we argue that the evolution of urban mobility is characterized by the presence of two co-existing, but orthogonal, sources of variability: temporal and spatial (Fig \ref{fig:contribution}).
Temporal variability can be viewed as the stochasticity characterizing all possible future scenarios of mobility demand over time, i.e. ``how will the need for transportation evolve in the near future?''.
On the other hand, spatial variability encodes the stochasticity of mobility demand in longitude-latitude space, i.e. ``how likely is it that someone will request for transportation in a specific area of the city?''.
In light of this, we believe predictive models for urban mobility should acknowledge the presence of these sources of variability, and the design of novel architectures should aim to effectively leverage this disentanglement.
Crucially, we argue that most efforts in the field currently fail to encode these sources of variability in two substantial ways: temporal variability is mostly ignored by the vast majority of deep learning architectures, which typically rely on some kind of recurrent neural architecture.
Because of its uniquely \emph{deterministic} hidden states, the only source of randomness in RNNs is found in the conditional output probability model, which has proven to be a limitation when attempting to model strong and complex dependencies among the output variables at different timesteps \cite{ChungKastnerEtAl2015}.
On the other hand, when modeling urban mobility in longitude-latitude space, spatial variability is typically approached through either (i) a spatial discretization (e.g. ConvLSTMs), or (ii) a Gaussian mixture model to describe the conditional output distribution.
We argue that both of these approaches could exhibit structural limitations when faced with the kind of spatial variability characterizing urban mobility densities (Fig \ref{fig:contribution2}).

\smallskip \noindent\textbf{Paper contributions} \hspace{4mm} The contributions of this paper are threefold.
First, we build on recent advances in deep generative models and propose \emph{recurrent flow networks} (RFN) to structurally encode both temporal and spatial variability in deep learning models for mobility demand prediction. 
On one hand, as have others in different domains \cite{ChungKastnerEtAl2015, FraccaroEtAl2016, KrishnanEtAl2016, KarlEtAl2017}, we bring together the representative power of RNNs with the consistent handling of (temporal) uncertainties given by probabilistic approaches.
On the other, we propose the use of Conditional Normalizing Flows (CNFs) \cite{WinklerEtAl2020} as a general approach to define arbitrarily expressive output probability distributions under temporal dynamics.
By doing so, the model can exploit the disentanglement in the variability of the demand generating process, which is advantageous when learning a model that is more effective and generalizable. 

Second, we show how the combination of deterministic and stochastic hidden states in RNNs, together with conditional normalizing flows enable predictive models to generate fine-grained urban mobility patterns.
Specifically, we focus on the problem of modeling the spatial distribution of transportation demand characterized by users of three real Mobility-on-Demand systems.

Third, this work highlights how, by leveraging its structural disentanglement of spatio-temporal variability, RFNs exhibit a series of desirable properties of fundamental practical importance for any system operator.
Specifically, results show an interesting ability of the model to (i) represent highly multimodal distributions over urban topologies, thus obtaining more reliable estimates compared to classic space discretization techniques or mixture models to describe the conditional output distribution, and (ii) generate reliable long-term predictions as a consequence of better encoding the temporal variability in mobility demand.

\section{Related Work}
\label{sec:related_work}
In this section, we review the literature that is relevant to our study in three main streams: (i) differences and similarities between two main classes of temporal models, i.e., dynamic Bayesian networks (DBNs) and recurrent neural networks (RNNs), (ii) the intersection of these two classes through recent recurrent latent variable models, and (iii) state-of-the-art deep learning architectures for the task of urban mobility modeling.

\subsection{Temporal models}
\label{subsec:temporal_models}
Historically, dynamic Bayesian networks (DBNs), such as hidden Markov models (HMMs) and state space models (SSMs) \cite{KoopmanEtAl2001}, have characterized a unifying probabilistic framework with illustrious successes in modeling time-dependent dynamics.
Advances in deep learning architectures however, shifted this supremacy towards the field of Recurrent Neural Networks (RNNs).
At a high level, both DBNs and RNNs can be framed as parametrizations of two core components: 1) a \emph{transition} function characterizing the time-dependent evolution of a learned internal representation, and 2) an \emph{emission} function denoting a mapping from representation space to observation space.

\noindent Despite their attractive probabilistic interpretation, the biggest limitation preventing the widespread application of DBNs in the deep learning community, is that inference can be exact only for models typically characterized by either simple transition/emission functions (e.g. linear Gaussian models) or relatively simple internal representations. 
On the other hand, RNNs are able to learn long-term dependencies by parametrizing a transition function of richly distributed deterministic hidden states. 
To do so, current RNNs typically rely on gated non-linearities such as long short-term memory (LSTMs) \cite{HochreiterSchmidhuber1997} cells and gated recurrent units (GRUs) \cite{ChungEtAl2014}, allowing the learned representation to act as internal memory for the model.

\subsection{Recurrent latent variable models}
\label{subsec:recurren_lv_models}
\noindent More recently, evidence has been gathered in favor of combinations bringing together the representative power of RNNs with the consistent handling of uncertainties given by probabilistic approaches \cite{ChungKastnerEtAl2015, FraccaroEtAl2016, KrishnanEtAl2016, KarlEtAl2017}.
The core concept underlying recent developments is the idea that, in current RNNs, the only source of variability is found in the conditional emission distribution (i.e. typically a unimodal distribution or a mixture of unimodal distributions), making  these models inappropriate when modeling highly structured data. 

A number of works have concentrated on building well-specified models for the prediction of spatio-temporal sequences.
One line of research particularly relevant for our work focuses on the definition of more flexible emission functions for sequential models \cite{RasulEtAl2021, KumarEtAl2020, CastrejonEtAl2019}.
As we also do in this paper, these works argue that simpler output models may turn out limiting when dealing with structured and potentially high dimensional data distributions (e.g. images, videos).
The performance of these models is highly dependent on the specific architecture defined in the conditional output distribution, as well as on how stochasticity is propagated in the transition function.
In VideoFlow \cite{KumarEtAl2020} and in \cite{RasulEtAl2021}, the authors similarly use normalizing flows to parametrize the emission function for the tasks of video generation and multi-variate time series forecasting, respectively. 
In VideoFlow, the latent states representing the temporal evolution of the system are defined by the conditional base distribution of a normalizing flow.
In \cite{RasulEtAl2021} the authors also propose to use conditional affine coupling layers in order to model the output variability. 
In \cite{CastrejonEtAl2019}, the authors address the task of video generation by defining a hierarchical version of the VRNN and a ConvLSTM decoder. 
While the latter also combines stochastic and deterministic states, as for the case of VideoFlow, the emission function is specifically focused on image modeling tasks.
However, to the author's best knowledge, there is no track of previous attempts using similar concepts for the task of urban mobility modeling. 

\subsection{Recurrent models for urban mobility modeling}
\label{subsec:recurrent_models_for_urban_mobility}
\smallskip \noindent Within the transportation domain, traditional approaches rely on spatial discretizations of the urban topology \cite{YuanEtAl2018, PetersenEtAl2019, WangEtAl2018}, which allow for the prediction of spatio-temporal sequences with discrete support.
Within this line of research, Convolutional LSTMs \cite{XingjianEtAl2015} represent a particularly successful neural architecture. 
Specifically, recurrence through convolutions is a natural fit for multi-step spatio-temporal prediction by taking advantage of the spatial invariance of discretized representations, thus requiring significantly fewer parameters.
Because of this, ConvLSTMs have been applied to a variety of different tasks both within and outside the transportation domain, such as precipitation nowcasting \cite{XingjianEtAl2015}, physics and video prediction \cite{FinnEtAl2016}, traffic accident prediction \cite{YuanEtAl2018}, bus arrival time forecasting \cite{PetersenEtAl2019} and travel demand prediction \cite{WangEtAl2018}.
In our work, we argue this line of literature to be not well-suited for the task of mobility demand prediction in several fundamental ways. 
First of all, ConvLSTMs are ultimately fully deterministic models, ignoring the possibility of stochasticity in the temporal transitions, thus incurring less accurate uncertainty estimates for the observed data.
Moreover, in order to exploit the spatial invariance of convolutions, ConvLSTMs require the data distribution to be described by discrete support (i.e., pixel values on a grid), thus enforcing for an unnecessary discretization of space in the task of geospatial transportation density estimation, which is naturally defined on continuous latitude-longitude space.
In this work, we argue that explicitly allowing for the propagation of uncertainty in the temporal transitions through the use of stochastic latent variables, as well as removing the necessity of a discretized output distribution by directly modeling observed data in latitude-longitude space, are fundamental properties for effective modeling of urban mobility data. 

\section{Background}
\label{sec:background}
In this section, we introduce the notation and theoretical foundations underlying our work in the context of RNNs and mixture density models (Section \ref{subsec:rnns}), stochastic RNNs (Section \ref{subsec:stochastic_rnns}) and normalizing flows for density estimation (Section \ref{subsec:normalizing flows}).

\subsection{RNNs and Mixture Density Outputs}
\label{subsec:rnns}
Recurrent neural networks are widely used to model variable-length sequences $\mathbf{x} = (\mathbf{x}_1, \mathbf{x}_2, \ldots, \mathbf{x}_T)$, possibly influenced by external covariates $\mathbf{u} = (\mathbf{u}_1, \mathbf{u}_2, \ldots, \mathbf{u}_T)$. 
The core assumption underlying these models is that all observations $\mathbf{x}_{1:t}$ up to time $t$ can be summarized by a learned deterministic representation $\mathbf{h}_t$. 
At any timestep $t$, an RNN recursively updates its hidden state $\mathbf{h}_t \in \mathbb{R}^p$ by computing:
\begin{equation}
    \mathbf{h}_t = f_{\theta_{\bh}}(\mathbf{u}_t, \mathbf{h}_{t-1}),
\end{equation}
where $f$ is a deterministic non-linear transition function with parameters $\theta_{\bh}$, such as an LSTM cell or a GRU.
The sequence is then modeled by defining a factorization of the joint probability distribution as the following product of conditional probabilities:
\begin{align}
    p(\mathbf{x}_1, \mathbf{x}_2, \ldots \mathbf{x}_T) &= \prod_{t=1}^{T}{p(\mathbf{x}_t|\mathbf{x}_{<t})} \nonumber\\
    p(\mathbf{x}_t|\mathbf{x}_{<t}) &= p(\mathbf{x}_t \mid \mathbf{h}_t) p(\mathbf{h}_t \mid \mathbf{h}_{t-1}, \mathbf{u}_{t}), \label{eq:rnn}
\end{align}
where $p(\mathbf{h}_t \mid \mathbf{h}_{t-1}, \mathbf{u}_{t}) = \delta(\mathbf{h}_t - \tilde{\mathbf{h}}_t)$, i.e. $\mathbf{h}_t$ follows a delta distribution centered in $\tilde{\mathbf{h}}_t = f_{\theta_{\bh}}(\mathbf{u}_t, \mathbf{h}_{t-1})$.

\smallskip \noindent When modeling complex distributions on real-valued sequences, a common choice is to represent the emission function with a \emph{mixture density network} (MDN), as in \cite{GravesEtAl2013}. 
The idea behind MDNs is to use the output of a neural network to parametrize a Gaussian mixture model. 
In the context of RNNs, a subset of the outputs at time $t$ is used to define the vector of mixture proportions $\boldsymbol{\pi}_t$, while the remaining outputs are used to define the means $\bmu_t$ and covariances $\bSigma_t$ for the corresponding mixture components. Under this framework, the probability of $\mathbf{x}_t$ is defined as follows:
\begin{equation}
p_{\boldsymbol{\pi}_t, \boldsymbol{\mu}_t, \boldsymbol{\Sigma}_t}(\mathbf{x}_t | \mathbf{x}_{<t}) = \sum_{k}^{K}{\pi_{k,t}\mathcal{N}(\mathbf{x}_t | \bmu_{k, t}, \bSigma_{k, t})},
\end{equation}
where $K$ is the assumed number of components characterizing the mixture.

\subsection{Stochastic Recurrent Neural Networks}
\label{subsec:stochastic_rnns}
As introduced in \cite{FraccaroEtAl2016}, a \emph{stochastic recurrent neural network} (SRNN) represents a specific architecture combining deterministic RNNs with fully stochastic state-space model layers. At a high level, SRNNs build a hierarchical internal representation by stacking a state-space model transition $f_{\theta_{\bz}}(\bz_{t-1}, \bh_t)$ on top of a RNN $f_{\theta_{\bh}}(\bh_{t-1}, \bu_t )$. The emission function is further defined by skip-connections mapping both deterministic ($\bh_t$) and stochastic ($\bz_t$) states to observation space ($\bx_t$). Assuming that the starting hidden states $\bh_0, \bz_0$ and inputs $\bu_{1:T}$ are given, the model is defined by the following factorization (where, for notational convenience, we drop the explicit conditioning of the joint distribution on $\bz_0$, $\bh_0$ and $\bu_{1:T}$): 
\begin{align}
    p(\bx, \bz, \bh) = \prod_{t=1}^{T}p_{\theta_{\bx}}(\bx_t | \bz_t, \bh_t) & p_{\theta_{\bz}}(\bz_t | \bz_{t-1}, \bh_t) p_{\theta_{\bh}}(\bh_t | \bh_{t-1}, \bu_t ),
    \label{eq:srnn}
\end{align}
where the emission and transition distributions have parameters $\theta_{\bx}, \theta_{\bz}, \theta_{\bh}$, and where we assume that $\mathbf{h}_{t}$ follows a delta distribution centered in $\mathbf{h}_{t} = f_{\theta_{\bh}}(\mathbf{h}_{t-1}, \mathbf{u}_t)$.

\subsection{Normalizing Flows for probabilistic modeling}
\label{subsec:normalizing flows}
Normalizing flows represent a flexible approach to define rich probability distributions over continuous random variables. 
At their core, flow-based models define a joint distribution over a $D$-dimensional vector $\bx$ by applying a transformation\footnote{Not to be confused with the time-horizon $T$ from e.g. Eq. \eqref{eq:rnn}. In general, the distinction should be clear from the context.} $T$ to a real vector $\bb$ sampled from a (usually simple) base distribution $p_{b}(\bb)$:
\begin{equation*}
\bx = T(\bb) \hspace{3mm} \text{where} \hspace{3mm} \bb \sim  p_{b}(\bb).
\end{equation*}

\noindent In order for the density of $\bx$ to be well-defined, some important properties need to be satisfied. In particular, the transformation $T$ must be \emph{invertible} and both $T$ and $T^{-1}$ must be \emph{differentiable}. Such a transformation is known as a \emph{diffeomorphism} (i.e. a differentiable invertible transformation with differentiable inverse). If these properties are satisfied, the model distribution on $\bx$ can be obtained by the change of variable formula: 
\begin{align}
    p_{\bx}(\bx) &= p_{b}(\bb)|\det J_T(\bb)|^{-1}\\
    \log\left(p_{\bx}(\bx)\right) &= \log\left(p_{b}(\bb)\right) + \log\left(|\det J_T(\bb)|^{-1}\right),
\end{align}
where $\bb = T^{-1}(\bx)$ and the Jacobian $J_T(\bb)$ is the $D \times D$ matrix of all partial derivatives of $T$. 
In practice, the transformation $T$ and the base distribution $p_b(\bb)$ can have parameters of their own (e.g. $p_b(\bb)$ could be a multivariate normal with mean and covariance also parametrized by any flexible function).
The fundamental property which makes normalizing flows so attractive, is that invertible and differentiable transformations are \emph{composable}. 
That is, given two transformations $T_1$ and $T_2$, their composition $T_2 \circ T_1$ is also invertible and differentiable, with inverse and Jacobian determinant given by:
\begin{align}
    (T_2 \circ T_1)^{-1} &= T_{1}^{-1} \circ T_{2}^{-1}\\
    \det J_{T_2 \circ T_1}(\bb) &=  \det J_{T_2}(T_1(\bb)) \cdot \det J_{T_1}(\bb).
\end{align}
As a result, this framework allows to construct arbitrarily complex transformations by composing multiple stages of simpler transformations, without sacrificing the ability of exactly calculating the (log) density $p_{\bx}(\bx)$.

\smallskip
In \cite{DinhEtAl2017}, the authors introduce a bijective function of particular interest for this paper. This transformation, known as an \emph{affine coupling layer}, exploits the simple observation that the determinant of a triangular matrix can be efficiently computed as the product of its diagonal terms. 
Concretely, given a $D$-dimensional sample from the base distribution $\bb$ and $d < D$, this property is exploited by defining the output $\bx$ of an affine coupling layer as follows:
\begin{align}
    \bx_{1:d} &= \bb_{1:d} \label{eq:affine_coupling_layer1}\\
    \bx_{d+1:D} &= \left(\bb_{d+1:D} - t(\bb_{1:d})\right) \odot \exp\left(-s(\bb_{1:d})\right),  \label{eq:affine_coupling_layer2}
\end{align}

where $s$ and $t$ are arbitrarily complex scale and translation functions from $ \mathbb{R}^d \mapsto \mathbb{R}^{D-d}$ and $\odot$ is the element-wise or Hadamard product. Since the forward computation defined in Eq. \eqref{eq:affine_coupling_layer1} and Eq. \eqref{eq:affine_coupling_layer2} leaves the first $d$ components unchanged, these transformations are usually combined by composing coupling layers in an alternating pattern, so that components unchanged in one layer are effectively updated in the next (for a more in-depth treatment of normalizing flows, the reader is referred to \cite{PapamakariosEtAl2021, KobyzevEtAl2021}).

\section{Recurrent Flow Networks}
\label{sec:rfn}

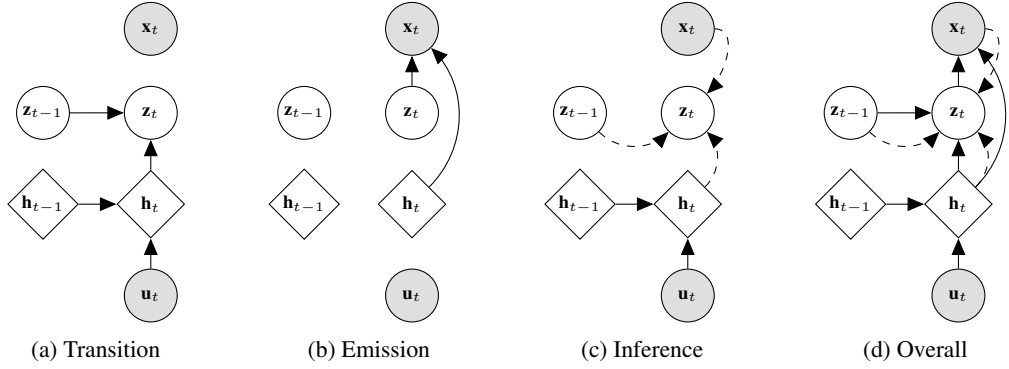
\begin{figure*}[t]
\centering
\begin{subfigure}{0.22\textwidth}
\centering
\begin{tikzpicture}[x=.3cm,y=.4cm]
\node[obs]    			   					(u_t)     {\scriptsize $\bu_{t}$} ;
\node[det, above=of u_t, xshift=0cm, yshift=0cm, minimum size=0.9cm]         (h_t)     {\scriptsize $\bh_{t}$} ; 
\node[latent, above=of h_t, xshift=0cm, yshift=0.0cm] (z_t) {\scriptsize $\bz_{t}$};
\node[obs, above=of z_t, xshift=0cm, yshift=0cm]   (x_t)     {\scriptsize $\bx_{t  }$} ; 
\node[det, left=of h_t, xshift=-0.2cm, yshift=0cm, minimum size=0.9cm]         (h_t_1)     {\scriptsize $\bh_{t-1}$} ; 
\node[latent, above=of h_t_1, xshift=0cm, yshift=-0.02cm] (z_t_1) {\scriptsize $\bz_{t-1}$};
\edge {u_t} {h_t} ; 
\edge {h_t} {z_t} ;
\edge {h_t_1} {h_t} ; 
\edge {z_t_1} {z_t} ; 
\end{tikzpicture}
\caption{Transition}
\end{subfigure}\hspace{\fill}
~
\begin{subfigure}{0.22\textwidth}
\centering
\begin{tikzpicture}[x=.3cm,y=.4cm]
\node[obs]    			   					(u_t)     {\scriptsize $\bu_{t}$} ;
\node[det, above=of u_t, xshift=0cm, yshift=0cm, minimum size=0.9cm]         (h_t)     {\scriptsize $\bh_{t}$} ; 
\node[latent, above=of h_t, xshift=0cm, yshift=0.0cm] (z_t) {\scriptsize $\bz_{t}$};
\node[obs, above=of z_t, xshift=0cm, yshift=0cm]   (x_t)     {\scriptsize $\bx_{t  }$} ; 
\node[det, left=of h_t, xshift=-0.2cm, yshift=0, minimum size=0.9cm]         (h_t_1)     {\scriptsize $\bh_{t-1}$} ; 
\node[latent, above=of h_t_1, xshift=0cm, yshift=-0.02cm] (z_t_1) {\scriptsize $\bz_{t-1}$};
\draw [->] (h_t) to [out=45,in=-45] (x_t);
\edge {z_t} {x_t} ;
\end{tikzpicture}
\caption{Emission}
\end{subfigure}\hspace{\fill}
~
\begin{subfigure}{0.22\textwidth}
\centering
\begin{tikzpicture}[x=.3cm,y=.4cm]
\node[obs]    			   					(u_t)     {\scriptsize $\bu_{t}$} ;
\node[det, above=of u_t, xshift=0cm, yshift=0cm, minimum size=0.9cm]         (h_t)     {\scriptsize $\bh_{t}$} ; 
\node[latent, above=of h_t, xshift=0cm, yshift=0.0cm] (z_t) {\scriptsize $\bz_{t}$};
\node[obs, above=of z_t, xshift=0cm, yshift=0cm]   (x_t)     {\scriptsize $\bx_{t  }$} ; 
\node[det, left=of h_t, xshift=-0.2cm, yshift=0cm, minimum size=0.9cm]         (h_t_1)     {\scriptsize $\bh_{t-1}$} ; 
\node[latent, above=of h_t_1, xshift=0cm, yshift=-0.02cm] (z_t_1) {\scriptsize $\bz_{t-1}$};
\edge {h_t_1} {h_t};
\edge {u_t} {h_t};
\draw [dashed, ->] (z_t_1) to [out=-45,in=-135] (z_t);
\draw [dashed, ->] (x_t) to [out=-0,in=45] (z_t);
\draw [dashed, ->] (h_t) to [out=45,in=-45] (z_t);
\end{tikzpicture}
\caption{Inference}
\end{subfigure}\hspace{\fill}
~
\begin{subfigure}{0.22\textwidth}
\centering
\begin{tikzpicture}[x=.3cm,y=.4cm]
\node[obs]    			   					(u_t)     {\scriptsize $\bu_{t}$} ;
\node[det, above=of u_t, xshift=0cm, yshift=0cm, minimum size=0.9cm]         (h_t)     {\scriptsize $\bh_{t}$} ; 
\node[latent, above=of h_t, xshift=0cm, yshift=0.0cm] (z_t) {\scriptsize $\bz_{t}$};
\node[obs, above=of z_t, xshift=0cm, yshift=0cm]   (x_t)     {\scriptsize $\bx_{t  }$} ; 
\node[det, left=of h_t, xshift=-0.2cm, yshift=0cm, minimum size=0.9cm]         (h_t_1)     {\scriptsize $\bh_{t-1}$} ; 
\node[latent, above=of h_t_1, xshift=0cm, yshift=-0.02cm] (z_t_1) {\scriptsize $\bz_{t-1}$};
\edge {u_t} {h_t} ; 
\edge {h_t} {z_t} ;
\edge {h_t_1} {h_t} ; 
\edge {z_t_1} {z_t} ; 
\draw [->] (h_t) to [out=45,in=-45] (x_t);
\edge {z_t} {x_t} ;
\draw [dashed, ->] (z_t_1) to [out=-45,in=-135] (z_t);
\draw [dashed, ->] (x_t) to [out=-0,in=45] (z_t);
\draw [dashed, ->] (h_t) to [out=45,in=-45] (z_t);
\end{tikzpicture}
\caption{Overall}
\end{subfigure}
\caption{Graphical model of the operations defining the RFN: a) transition function defined in Eq. \eqref{eq:rfn_t_1} and Eq. \eqref{eq:rfn_t_2}; b) emission function as in Eq. \eqref{eq:rfn_prior} and Eq. \eqref{eq:rfn_flow}; c) inference network using Eq. \eqref{eq:rfn_inf}; d) overall RFN graphical model. Shaded nodes represent observed variables, while un-shaded nodes represent either deterministic (diamond-shaped) or stochastic (circles) hidden states. For sequence generation, a traditional approach is to use $\bu_t = \bx_{t-1}$.}
\label{fig:rfn}
\end{figure*}

\begin{figure*}[t]
\centering
\begin{subfigure}{0.3\textwidth}
\centering
\includegraphics[width=\columnwidth]{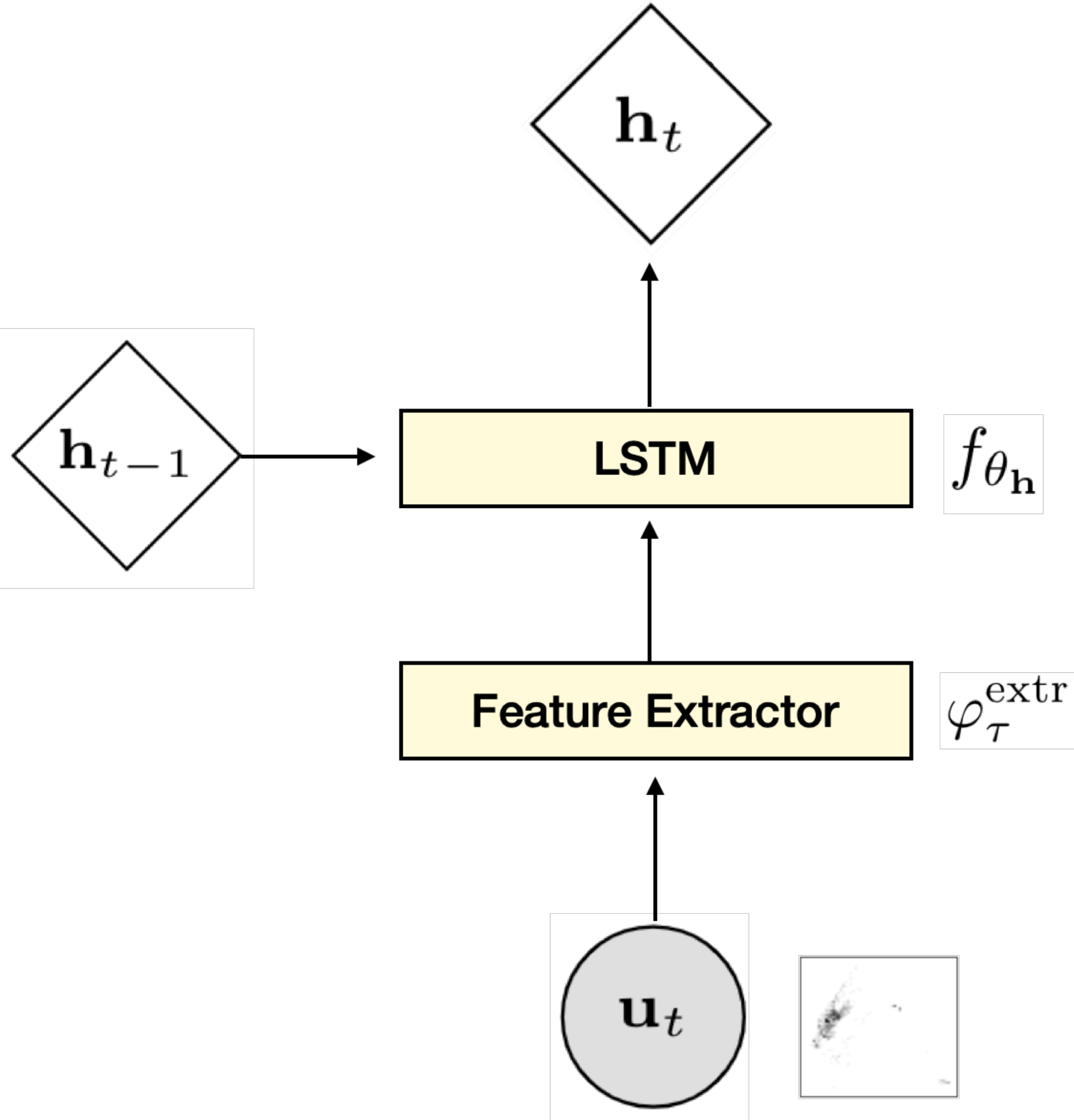}
\caption{RNN Transition}
\label{fig_rev:rfn_diagram_a}
\end{subfigure}\hspace{\fill}
~
\begin{subfigure}{0.3\textwidth}
\centering
\includegraphics[width=1.32\columnwidth,]{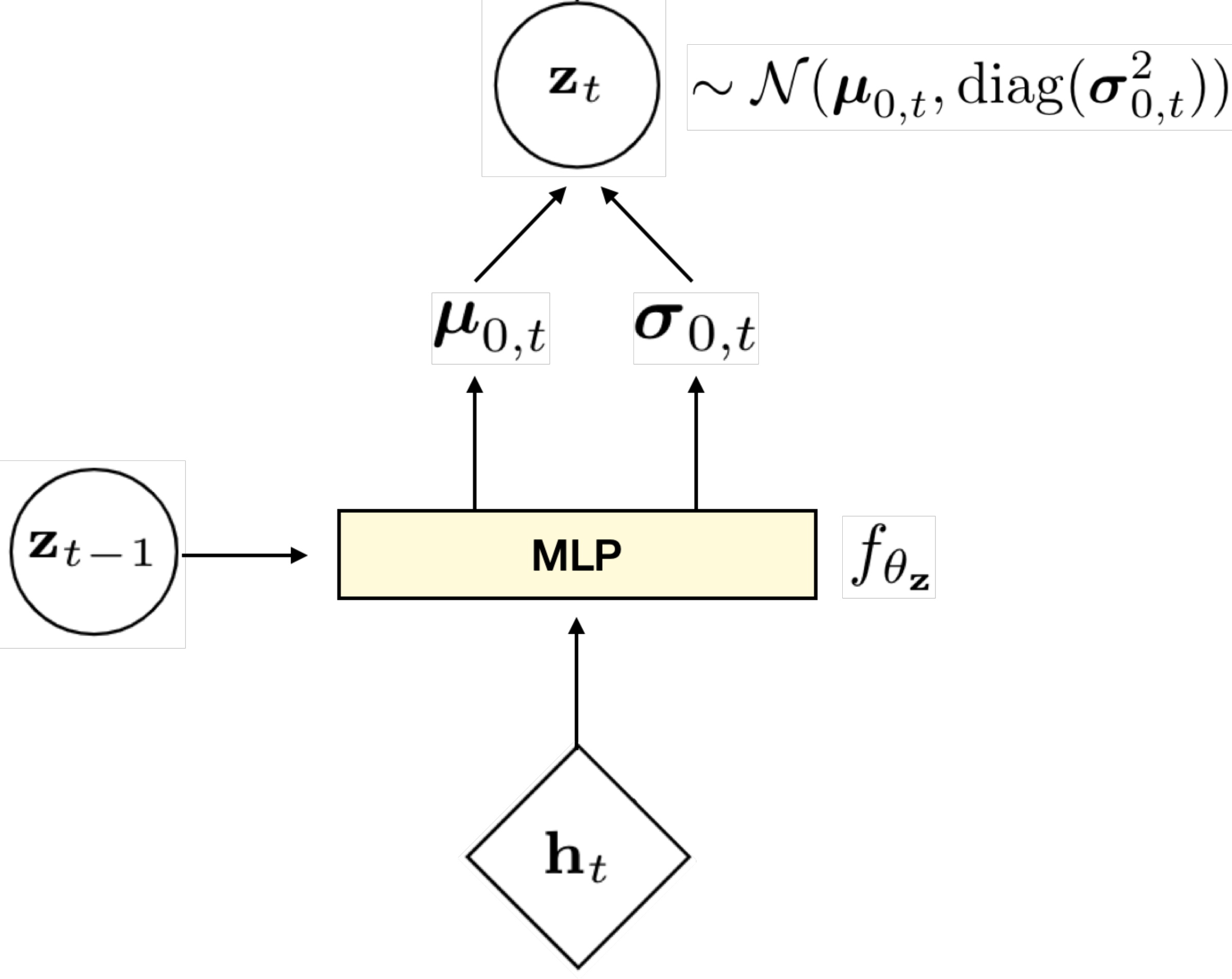}
\caption{Conditional Prior}
\label{fig_rev:rfn_diagram_b}
\end{subfigure}\hspace{\fill}
~
\begin{subfigure}{0.3\textwidth}
\centering
\includegraphics[width=0.79\columnwidth,]{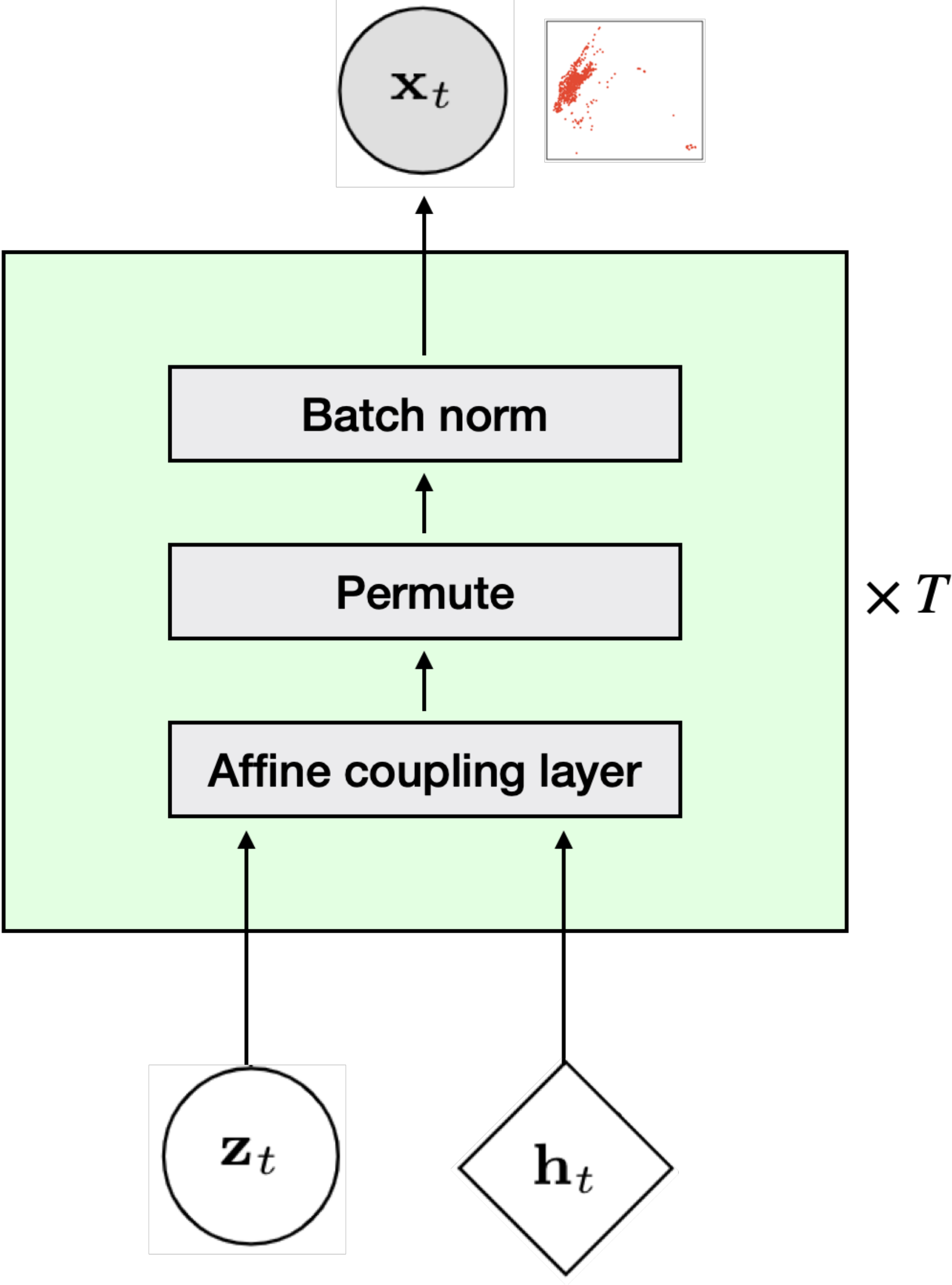}
\caption{Conditional Output}
\label{fig_rev:rfn_diagram_c}
\end{subfigure}
\caption{Architecture diagram of the learnable components of the RFN: (a) deterministic transition function (Eq. \ref{eq:rfn_t_1}), (b) conditional prior distribution over the latent variable $\bz_t$ (Eq. \ref{eq:rfn_t_2}), and (c) modules of the conditional normalizing flow used to parametrize the output distribution (Eq. \ref{eq:rfn_prior} - \ref{eq:rfn_flow}).}
\label{fig:rfn_diagram}
\end{figure*}

In this section, we define the generative model $p_{\theta}$ and inference network $q_{\phi}$ characterizing the RFN for the purpose of sequence modeling. RFNs explicitly model temporal dependencies by combining deterministic and stochastic layers. The resulting intractability of the posterior distribution over the latent states $\bz_{1:T}$, as in the case of VAEs \cite{KingmaEtAl2014, RezendeEtAl2014}, is further approached by learning a tractable approximation through \emph{amortized variational inference}.
The schematic view of the RFN is shown in Fig \ref{fig:rfn} (probabilistic graphical model), and Fig \ref{fig:rfn_diagram} (architectural diagram).

\smallskip \noindent \textbf{Generative model} \hspace{4mm} As in \cite{FraccaroEtAl2016}, the transition function of the RFN interlocks a state-space model with an RNN:
\begin{align}
    &\bh_t = f_{\theta_{\bh}}(\bh_{t-1}, \varphi^{\text{extr}}_{\tau}(\bu_t)) \label{eq:rfn_t_1}\\
    & \bz_t \sim \mathcal{N}(\bmu_{0, t}, \diag(\bsigma_{0, t}^2)), \label{eq:rfn_t_2} \\ 
    & \text{with } [\bmu_{0, t}, \bsigma_{0, t}] = f_{\theta_{\bz}}(\bz_{t-1}, \bh_t), \nonumber
\end{align}
where $\bmu_{0, t}$ and $\bsigma_{0, t}$ represent the parameters of the conditional prior distribution over the stochastic hidden states $\bz_{1:T}$. In our implementation, $f_{\theta_{\bh}}$ and $f_{\theta_{\bz}}$ are respectively an LSTM cell and a feed-forward neural network, with parameters $\theta_{\bh}$ and $\theta_{\bz}$.
In Eq. \eqref{eq:rfn_t_1}, $\varphi^{\text{extr}}_{\tau}$ can also be a neural network extracting features from $\bu_t$.
Unlike the SRNN, the learned representations (i.e. $\bz_{1:T}$, $\bh_{1:T}$) are used as conditioners for a CNF parametrizing the output distribution. 
That is, for every time-step $t$, we learn a complex distribution $p(\bx_t | \bz_t, \bh_t)$ by defining the conditional base distribution $p(\bb_t | \bz_t, \bh_t)$ and conditional coupling layers for the transformation $T_{\psi}$ as follows:
\begin{align}
    & \text{Conditional Prior:} \nonumber \\
    & \hspace{10mm} \bb_t \sim \mathcal{N}(\bmu_{b, t}, \diag(\bsigma_{b, t}^2)), \label{eq:rfn_prior}\\ 
    & \hspace{10mm} \text{with } [\bmu_{b, t}, \bsigma_{b, t}] = f_{\psi}(\bz_t, \bh_t) \nonumber\\\nonumber\\
    & \text{Conditional Coupling:} \nonumber \\
    & \hspace{10mm} \bx_{t, d+1:D} = \left(\bb_{t,d+1:D} - t_{\psi}(\bb_{t, 1:d}, \bz_{t}, \bh_t)\right) \odot \nonumber\\ 
    & \hspace{26mm} \exp\left(-s_{\psi}(\bb_{t, 1:d}, \bz_{t}, \bh_t)\right) \label{eq:rfn_flow}\\
    & \hspace{10mm} \bx_{t, 1:d} = \bb_{t, 1:d} \nonumber,
\end{align}
where $\bmu_{b, t}$ and $\bsigma_{b, t}$ represent the parameters of the conditional base distribution (determined by a learnable function $f_{\psi}$), while $s_{\psi}$ and $t_{\psi}$ denote the conditional scale and translation functions characterizing the coupling layers in the CNF. In our implementation, $f_{\psi}$, $s_{\psi}$ and $t_{\psi}$ are parametrized by neural networks. Together, Eq. \eqref{eq:rfn_prior} and Eq. \eqref{eq:rfn_flow} define the emission function $g_{\theta_{\bx}}(\bz_t, \bh_t)$, enabling the generative model to result in the factorization in Eq. \eqref{eq:srnn}.

\smallskip \noindent\textbf{Inference} \hspace{4mm} To perform inference it is sufficient to reason using the marginal likelihood of a probabilistic model. 
Consider a general probabilistic model with observations $\bx$, latent variables $\bz$ and parameters $\boldsymbol{\theta}$.
Variational inference introduces an approximate posterior distribution for the latent variables $q_{\theta}(\bz \mid \bx)$.
This formulation can be easily extended to posterior inference over the parameters $\boldsymbol{\theta}$, but in this work we will focus on inference over the latent variables only.
By following the variational principle \cite{JordanEtAl1999}, we obtain a lower bound on the marginal likelihood, often referred to as the negative free energy or evidence lower bound (ELBO):
\begin{align}
     & \log p_{\theta}(\bx) = \log \int p_{\theta}(\bx  |  \bz) p(\bz) d\bz \nonumber \\
     & = \log \int \frac{q_{\phi}(\bz  |  \bx)}{q_{\phi}(\bz | \bx)} p_{\theta}(\bx | \bz) p(\bz) d\bz \label{eq:elbo_general}\\
    & \geq -\mathbb{KL} \left[q_{\phi}(\mathbf{z} | \mathbf{x}) || \, p_{\theta}(\mathbf{z})\right] + \mathbb{E}_q \left[\log p_{\theta}(\bx | \bz) \right], \nonumber
\end{align}
where we use Jensen's inequality to obtain the final equation.
The bound consists of two terms: the first is the KL divergence between the approximate posterior and the prior distribution over latent variables $\bz$, thus acting as a regularizer, and the second represents the reconstruction error.
Denoting $\theta$ and $\phi$ as the set of model and variational parameters respectively, variational inference offers a scheme for jointly optimizing parameters $\theta$ and computing an approximation to the posterior distribution.

\noindent In this work, we represent the approximate posterior distribution $q_{\phi}$ through an inference network that learns an inverse map from observations to latent variables.
The benefit of using an inference network lies in avoiding the need to compute per-sample variational parameters, but instead learn a global set of parameters $\boldsymbol{\phi}$ valid for inference at both training and test time.
Crucially, this allows to \emph{amortize} the cost of inference through the network's ability to generalize between posterior estimates for all latent variables.

\smallskip \noindent In the context of the RFN, the variational approximation directly depends on $\bz_{t-1}$, $\bh_t$ and $\bx_t$ as follows:
\begin{align}
    & q_{\phi}(\bz_t | \bz_{t-1}, \bh_t, \bx_t) = \mathcal{N}(\bmu_{z, t}, \diag(\bsigma_{z, t}^2)), \hspace{3mm} \label{eq:rfn_inf} \\ 
    & \text{with } [\bmu_{z, t}, \bsigma_{z, t}] = \varphi^{\text{enc}}_{\tau}(\bz_{t-1}, \bh_{t}, \bx_t) \nonumber,
\end{align}
where $\varphi^{\text{enc}}_{\tau}$  is an encoder network defining the parameters of the approximate posterior distribution  $\bmu_{z, t}$ and $\bsigma_{z, t}$. Given the above structure, the generative and inference models are tied through the RNN hidden state $\bh_t$, resulting in the factorization given by:
\begin{align}
    q_{\phi}(\bz_{1:T} | \bx_{1:T}) = \prod_{t=1}^{T} q_{\phi}(\bz_{t} | \bz_{t-1}, \bh_{t}, \bx_t).
\end{align}
In addition to the explicit dependence of the approximate posterior on $\bx_t$ and $\bh_t$, the inference network defined in Eq. \eqref{eq:rfn_inf} also exhibits an implicit dependence on $\bx_{1:t}$ and $\bh_{1:t}$ through $\bz_{t-1}$. This implicit dependency on all information from the past can be considered as resembling a \emph{filtering} approach from the state-space model literature \cite{KoopmanEtAl2001}. 
Concretely, we jointly optimize parameters $\theta$ and $\phi$ to compute an approximation to the unknown posterior distribution by maximizing the following step-wise evidence lower bound\footnote{Please refer to the Appendix for the derivation} (i.e. ELBO), through Algorithm \ref{alg:rfn}:
\begin{align}
     \mathcal{L}(\theta, \phi) &= \mathbb{E}_{q_{\phi}(\mathbf{z}_{1:T} | \mathbf{x}_{1:T})} \bigg[\sum_{t=1}^{T} \log p_{\theta}(\mathbf{x}_{t} | \mathbf{z}_{t}, \mathbf{h}_{t}) + \nonumber \\
     & \hspace{5mm} + \log p_{\theta}(\mathbf{h}_{t} | \mathbf{h}_{t-1}, \mathbf{u}_{t}) \bigg] + \label{eq:elbo}\\
    & - \sum_{t=1}^{T} \mathbb{KL} \left(q_{\phi}(\mathbf{z}_{t} | \mathbf{z}_{t-1}, \mathbf{h}_{t}, \mathbf{x}_{t}) || p_{\theta}(\mathbf{z}_{t} | \mathbf{z}_{t-1}, \mathbf{h}_{t}) \right). \nonumber
\end{align}

\begin{algorithm}[h]
\caption{Recurrent Flow Network training scheme}\label{alg:rfn}
\begin{algorithmic}[1]
\Require $\bX$: time-series of raw GPS coordinates
\Require $k$: histogram width/height dimension
\Require $\alpha$: step size hyperparameter
\State pre-process $\bX$ into $k \times k$ histograms, $\bU$ (Section \ref{subsec:spatial_representation})
\State randomly initialize $\theta, \phi$
\While{convergence criteria not met}
    \State Sample batch of time-series $\bx_{1:T} \in \bX, \bu_{1:T} \in \bU$
    \For{t}
        \State Evaluate $\nabla_{(\theta, \phi)} \mathcal{L}(\theta, \phi)$ as in Eq \ref{eq:elbo} w.r.t $\bx_{t}, \bu_{t}$
        \State Update parameters through gradient ascent: \\
        \hspace{1cm} $\theta^{'} \leftarrow \theta + \alpha\nabla_{\theta}\mathcal{L}(\theta, \phi)$\\
        \hspace{1cm} $\phi^{'} \leftarrow \phi + \alpha\nabla_{\phi}\mathcal{L}(\theta, \phi)$
    \EndFor
\EndWhile
\end{algorithmic}
\end{algorithm}

\section{Experiments}
\label{sec:experiments}
In this section, we present simulation results that demonstrate the performance of our proposed approach on different real-world scenarios.

\smallskip \noindent Concretely, we evaluate the proposed RFN on three transportation datasets:
\begin{itemize}
    \item \textbf{NYC Taxi (NYC-P/D)}: This dataset is released by the New York City Taxi and Limousine Commission. We focused on aggregating the taxi demand in 2-hour bins for the month of March 2016 containing 249,637 trip geo-coordinates. We further differentiated the task of modeling pick-ups (\mbox{i.e.} where the demand is) and drop-offs (\mbox{i.e.} where people want to go). In what follows, we denote the two datasets as NYC-P and NYC-D respectively.
    \item \textbf{Copenhagen Bike-Share (CPH-BS)}: This dataset contains geo-coordinates from users accessing the smartphone app of Donkey Republic, one of the major bike sharing services in Copenhagen, Denmark. As for the case of New York, we aggregated the geo-coordinates in 2-hour bins for the month of August, resulting in 87,740 app accesses.  
\end{itemize}

\smallskip \noindent For both New York and Copenhagen experiments we process the data so to discard corrupted geo-coordinates outside the area of interest. For the taxi experiments, we discarded coordinates related to trips either shorter than $30s$ or longer than $3h$, while in the bike-sharing dataset, we ensured to keep only one app access from the same user in a window of $5$ minutes. In both cases we divide the data temporally into train/validation/test splits using a ratio of $0.5/0.25/0.25$.

\subsection{Training}
We train each model using stochastic gradient ascent on the evidence lower bound $\mathcal{L}(\theta, \phi)$ defined in \mbox{Eq.} \eqref{eq:elbo} using the Adam optimizer \cite{KingmaEtAl2014}, with a starting learning rate of $0.003$ being reduced by a factor of $0.1$ every $100$ epochs without loss improvement (in our implementation, we used the \emph{ReduceLROnPlateau} scheduler in PyTorch with patience=100).
As in \cite{SnderbyEtAl2016}, we found that annealing the KL term in Eq. \eqref{eq:elbo} (using a scalar multiplier linearly increasing from 0 to 1 over the course of training) yielded better results. The final model was selected with an early-stopping procedure based on the validation performance.

\subsection{Models}
We compare the proposed RFN against various baselines assuming both continuous and discrete support for the output distribution. In particular, in the continuous case (i.e. where we assume to be modeling a 2-dimensional distribution directly in longitude-latitude space), we consider RNN, VRNN \cite{ChungKastnerEtAl2015} and SRNN \cite{FraccaroEtAl2016} models each using two different MDN-based emission distributions.
That is, we compare against a GMM output parametrized by Gaussians with either diagonal (MDN-Diag) or full (MDN-Full) covariance matrix. 
Moreover, in order to quantify the importance of modelling temporal variability we also implement an ablation of the RFN, inspired by \cite{RasulEtAl2021}, where the only difference lies in the deterministic transition function (in what follows, RNN-Flow).
On the other hand, when assuming discrete support for the output distribution (i.e. we
divide the map into tiled non-overlapping patches and view the pixels inside a patch as its measurements), we consider Convolutional LSTMs (ConvLSTM), \cite{XingjianEtAl2015} which leverage the spatial information encoded on the sequences by substituting the matrix operations in the standard LSTM formulation with convolutions\footnote{Please refer to the Appendix for complete experimental details}. 

\begin{figure}[t]
\centering
\begin{subfigure}{0.45\textwidth}
\centering
\includegraphics[width=1\columnwidth]{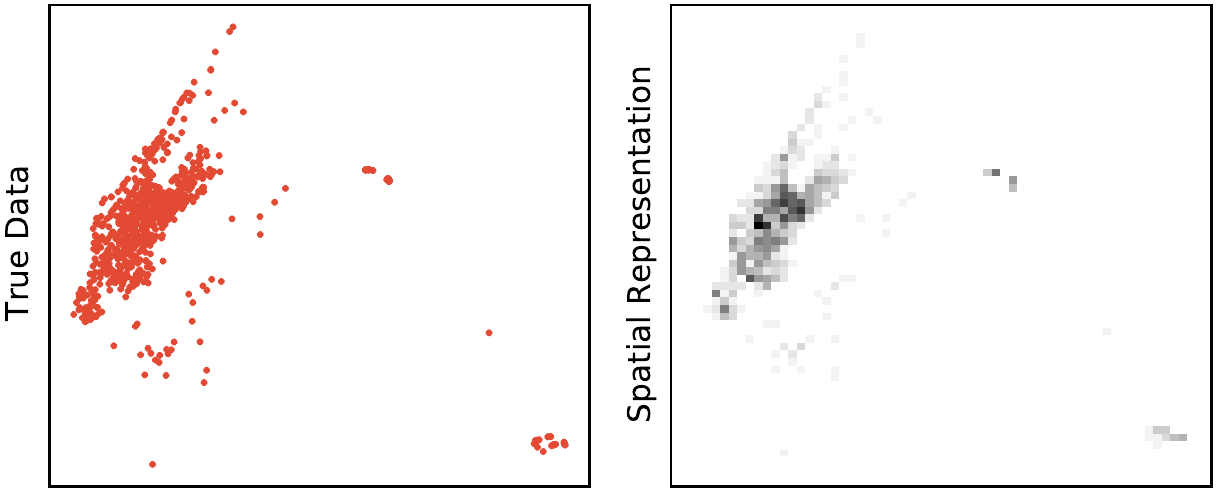}
\caption{10:00am}
\end{subfigure}
~
\begin{subfigure}{0.45\textwidth}
\centering
\includegraphics[width=1\columnwidth]{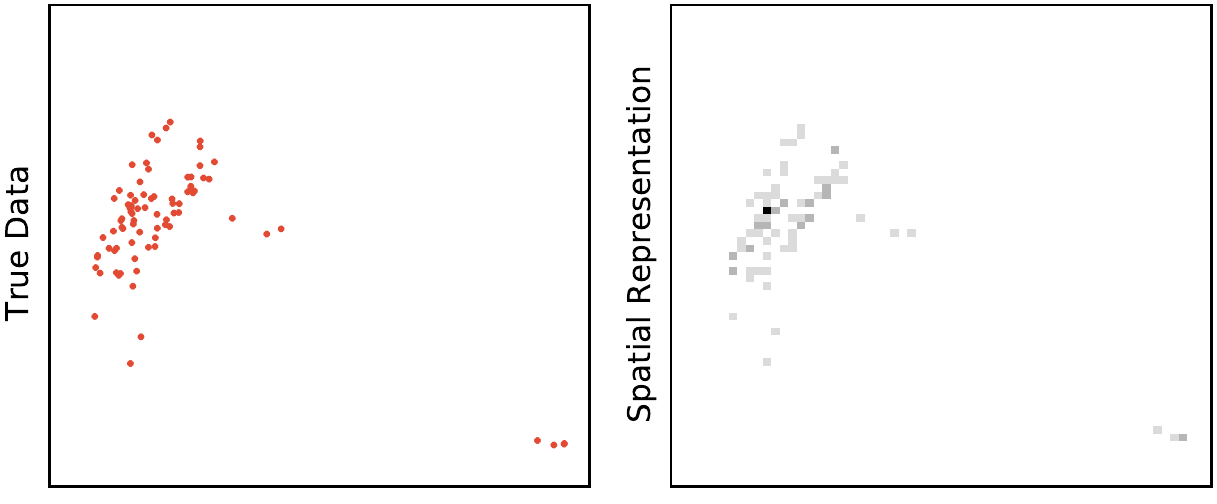}
\caption{02:00am}
\end{subfigure}
~
\begin{subfigure}{0.45\textwidth}
\centering
\includegraphics[width=1\columnwidth]{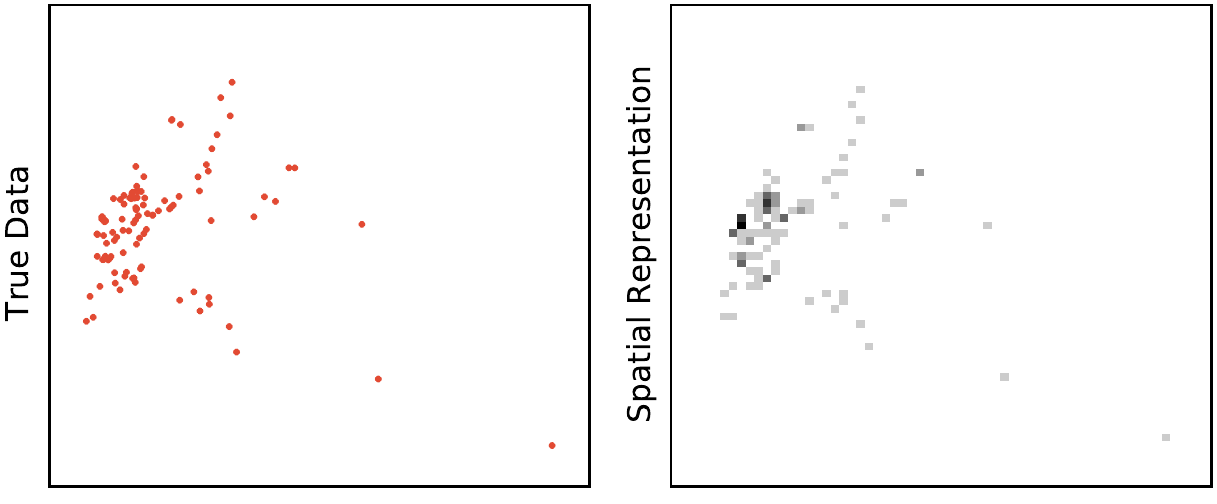}
\caption{07:00am}
\end{subfigure}

\caption{Visualization of the spatial representation for three distinct timesteps from the NYC-P dataset.}
\label{fig:approx}
\end{figure}

\subsection{Spatial representation}
\label{subsec:spatial_representation}
For the task of spatio-temporal density estimation, $\bx_{1:T}$ takes the form of a set of variable-length samples from the target distribution $p(\bx_{1:T})$. 
That is, for every time-step $t$, $\bx_{t}$ is a vector of geo-coordinates representing a corresponding number of taxi trips (NYC-P/D) or smartphone app accesses (CPH-BS).
We propose to process the data into a representation enabling the models to effectively handle data in a single batch computation.
As shown in Fig. \ref{fig:approx}, we choose to represent $\bx_t$ as a $k \times k$ normalized 2-dimensional histogram (in our implementation we set $k=64$). 
Given its ability to preserve the spatial structure of the data, we believe this representation to be well suited for spatio-temporal density estimation tasks.

\smallskip \noindent More precisely, the proposed representation is obtained by applying the following three-step procedure: 1) select data $\bx_t$, 2) build a 2-dimensional histogram computing the counts $c_{ij}, i, j = 1, \ldots k$ of the geo-coordinates falling in every cell of the $k \times k$ grid and 3) normalize the histogram such that $\sum_{i,j} c_{ij} = 1$.
By fixing $\bu_t = \bx_{t-1}$, this enables the definition of a sequence generation problem over spatial densities.
In practice, we found the above spatial representation to be both practical in dealing with variable-length geo-coordinate vectors, as well as effective, yielding better results. 
To the authors' best knowledge, this spatial approximation of the target distribution has never been used for the task of continuous spatio-temporal density modeling.
\begin{table}[t]
\normalsize
\centering
\begin{tabular}{l c c c}
    \hline
    Models & NYC-P & NYC-D & CPH-BS \\ [0.2ex] 
    \hline
    RNN-MDN-Diag & $163582 (\pm 492)$ & $143765 (\pm 376)$ & $49124 (\pm 210)$ \\ [0.15ex]
    RNN-MDN-Full & $164016 (\pm 501)$ & $146676 (\pm 391)$ & $50109 (\pm 223)$ \\[0.15ex]
    RNN-Flow & $165294 (\pm 578)$ & $146037 (\pm 387)$ & $49515 (\pm 291)$ \\[0.15ex]
    VRNN-MDN-Diag $\approx$ & $ 161345 (\pm 476)$ & $139964 (\pm 354)$ & $49231 (\pm 301)$ \\ [0.1ex]
    VRNN-MDN-Full $\approx$ & $162549 (\pm 432)$ & $143671 (\pm 368)$ & $49664 (\pm 243)$ \\[0.15ex]
    SRNN-MDN-Diag $\approx$& $164830 (\pm 549)$ & $143719 (\pm 342)$ & $49331 (\pm 276)$ \\[0.15ex]
    SRNN-MDN-Full $\approx$ & $164976 (\pm 465)$ & $147400 (\pm 393)$ & $49810 (\pm 287)$ \\[0.15ex]
    \textbf{RFN} $\boldsymbol{\approx}$ & $\mathbf{168734 (\pm 497)}$ & $\mathbf{148291 (\pm 407)}$ & $\mathbf{51100 (\pm 303)}$ \\
    \hline
    \end{tabular}%
    \caption{Average test log-likelihood (and standard deviations) for each task under the continuous support assumption. For the non-deterministic models (VRNN, SRNN, RFN) the approximation on the marginal log-likelihood is given with the $\approx$ sign.}
  \label{tab:results_continuous}%
\end{table}

\subsection{Results}
\label{subsec:results}
The goal of our experiments is to answer the following questions: (1) Can we learn fine-grained distributions of transportation demand on real-world urban mobility scenarios? (2) Given historical GPS traces of user movements, what are the advantages of using normalizing flows to characterize the conditional output distribution? (3) What are the advantages of explicitly representing uncertainty in the temporal evolution of urban mobility demand? 
\subsubsection{One-step Prediction}
\label{subsubsec:one_step_prediction}
In Table~\ref{tab:results_continuous} we compare test log-likelihoods on the tasks of continuous spatio-temporal demand modeling for the cases of New York and Copenhagen. 
We report exact log-likelihoods for both RNN-MDN-Diag and RNN-MDN-Full, while in the case of VRNNs, SRNNs and RFNs we report the importance sampling approximation to the marginal log-likelihood using 30 samples, as in \cite{RezendeEtAl2014}.
We see from Table 1 that RFN outperformes competing methods yielding higher log-likelihood across all tasks. The results support our claim that more flexible output distributions are advantageous when modeling potentially complex and structured temporal data distributions.
Table 1 also highlights how normalizing flows alone, i.e., modelling spatial variability and ignoring temporal variability in RNN-Flow, can be a bottleneck for effective modelling of fine-grained urban mobility patterns.
Crucially, in order to make full use of their output flexibility, we believe that predictive models must be equipped with the ability of modelling \emph{multiple possible futures}.
In this context, latent variables allow for an additional degree of freedom (i.e., different values of $\bz_t$ correspond to different future evolutions of mobility demand), on the other hand, deterministic transition functions - such as the ones present in RNN-based architectures - will necessarily converge to predicting an \emph{average of all futures}, thus resulting in more blurry and less-fine-grained spatial densities.

\begin{figure*}[t!]
\centering
\includegraphics[scale=1.1]{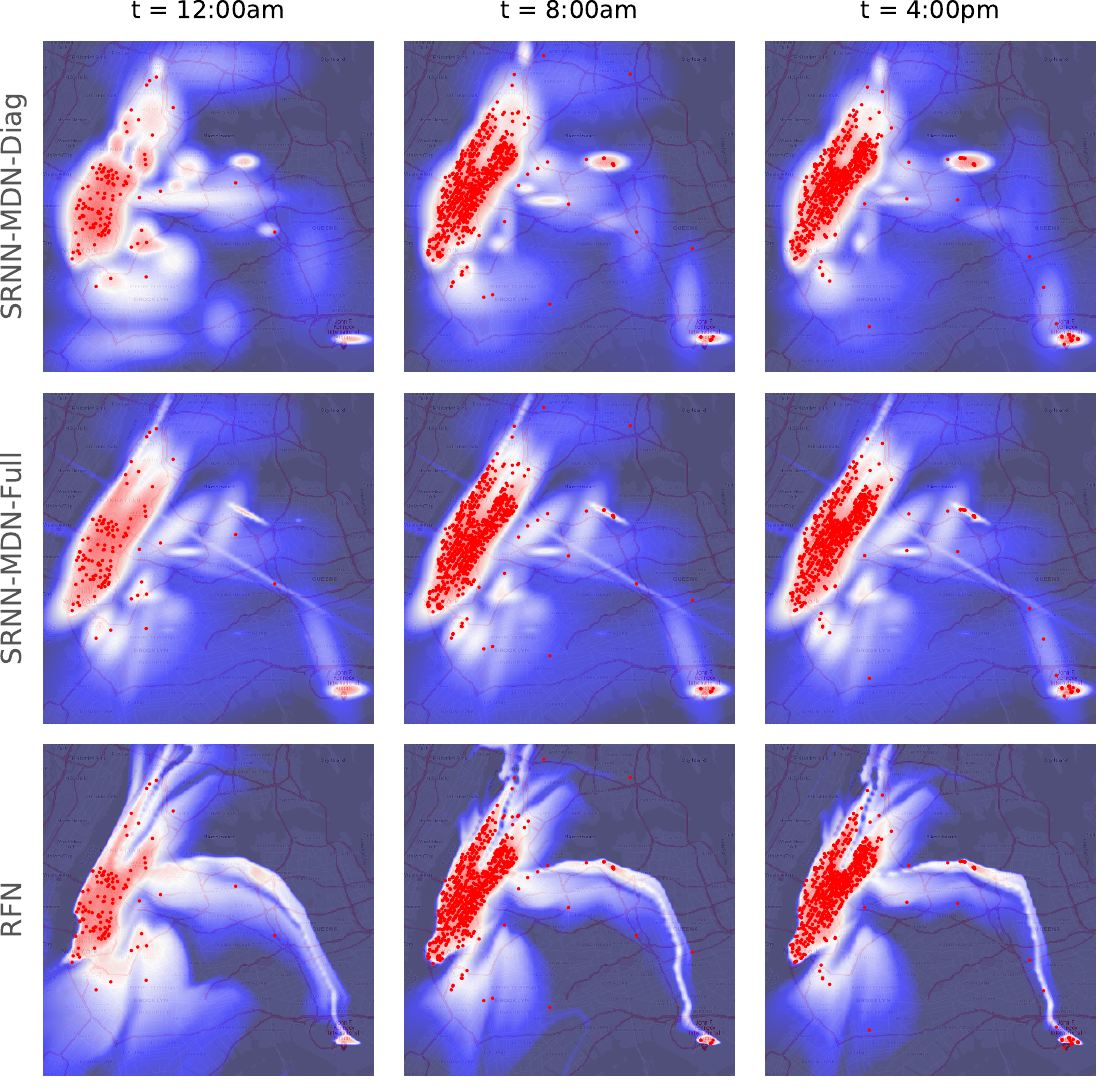}
\caption{Generated spatio-temporal densities from SRNN-MDN-Diag, SRNN-MDN-Full and RFN on the NYC-P dataset. The blue (low) to red (high) log-likelihood heatmaps show models defined by increasing flexibility (best viewed in color).}
\label{fig:grid}
\end{figure*}

\smallskip \noindent To further illustrate this, in Fig. \ref{fig:grid}, we show a visualization of the predicted spatial densities (one-step-ahead) from three of the implemented models at specific times of the day.
The heatmap was generated by computing the approximation of the marginal log-likelihood, under the respective model, on a $110 \times 110$ grid within the considered geographical boundaries.
The final plot is further obtained by mapping the computed log-likelihoods back into latitude-longitude space. 
Opposed to GMM-based densities, the figures show how the RFN exploits the flexibility of conditional normalizing flows to generate sharper distributions capable of better approximating complex shapes such as geographical landforms or urban topologies (e.g. Central Park or the sharper edges in proximity of the Hudson river along the west side of Manhattan). 

\subsubsection{Multi-step Prediction}
\label{subsubsec:multi_step_prediction}
In order to take reliable strategic decisions, service providers might also be interested in obtaining full roll-outs of demand predictions, opposed to 1-step predictions. To do so, we generate entire sequences in an autoregressive way (i.e., the prediction at timestep $t$ is fed back into the model at $t+1$) and analyze the ability of the proposed model to unroll for different forecasting horizons. From a methodological point of view, we are interested in measuring the effect of explicitly modeling the stochasticity in the temporal evolution of demand opposed to fully-deterministic architectures. To this regard, in Table \ref{tab:results_rollout} we compare the RFN with the most competitive deterministic benchmark (i.e. RNN-MDN-Full). As the results  suggest, the stochasticity in the transition probability allows the RFN to better capture the temporal dynamics, thus resulting in lower performance decay in comparison with the fully deterministic RNN-MDN assuming full covariance.

\begin{table}[t]
\normalsize
\centering
\begin{tabular}{l c c c c}
    \hline
    Models & t+2 & t+5 & t+10 & full (t+90) \\ [0.2ex] 
    \hline
    RNN-MDN & 162891 & 161065 & 160099 & 158922\\ [0.15ex]
    \textbf{RFN} $\boldsymbol{\approx}$ & \textbf{167509} & \textbf{167400} & \textbf{167359} & \textbf{167392} \\[0.15ex]
    \hline 
    \end{tabular}%
    \caption{Test log-likelihood comparison of RFN and RNN-MDN-Full for different forecast horizons on the NYC-P task.}
  \label{tab:results_rollout}%
\end{table}

\subsubsection{Quantization}
\label{subsubsec:quantization}
As a further analysis, we compare the proposed RFN with a Convolutional LSTM, under the assumption that the spatial map has been discretized in a $64 \times 64$ pixel space described by a Categorical distribution.  
This comparison is particularly relevant given the prevalence of ConvLSTMs in spatio-temporal travel modeling applications \cite{PetersenEtAl2019, YuanEtAl2018, WangEtAl2018}.
As previously introduced, the RFN is naturally defined by a continuous output distribution (in practice parametrized as a normalizing flow), thus, in order to characterize a valid comparison, we apply a quantization procedure to obtain a discrete output distribution for the RFN.
In particular, the implemented quantization procedure can be summarized with the following steps: (i) as in the continuous case, evaluate the approximated marginal log-likelihood under the trained RFN at the pixel-centers of a $64 \times 64$ grid, (ii) normalize the computed log-likelihood logits through the use of a softmax function and (iii) evaluate the log-likelihood under a Categorical distribution characterized by the probabilities computed in (ii), thus having values comparable with the output of the ConvLSTM.

\smallskip \noindent Table \ref{tab:results_discrete} compares test log-likelihoods on the task of discrete spatio-temporal demand modeling. 
To this regard, when considering results in Table \ref{tab:results_discrete}, two relevant observations must be underlined. 
First of all, the true output of the RFN (i.e. before quantization) is a continuous density, thus, its discretization will, by definition, result in a loss of information and granularity.
Secondly, and most importantly, the quantization is applied as post-processing evaluation step, thus, opposed to the implemented ConvLSTMs, the RFNs are not directly optimizing for the objective evaluated in Table \ref{tab:results_discrete}.

\noindent In light of this, the results under the discretized space assumption support even more our claims on the effectiveness of the RFN to approximate spatially complex distributions. 
Moreover, the ability of the RFN to model a continuous spatial density, opposed to a discretized approach as in the case of ConvLSTMs, has several theoretical and practical advantages.
For instance, RFNs are able to evaluate the log-likelihood of individual data points for anomaly and hotspot detection. Secondly, ConvLSTMs define a discretized space whose cells might have different natural landscape characteristics (e.g. rivers, lakes), thus effectively changing the dimension of the support in each bin and making comparisons of log-likelihoods across pixels an ill-posed question.
Furthermore, if for discrete output distributions exploring different levels of discretization would require to repeatedly train independent ConvLSTM networks, the post-processing quantization of the RFN allows discretization to be done instantaneously, thus enabling for fast prototyping and exploration of discretization levels.

\begin{table}[t]
\normalsize
\centering
\begin{tabular}{l c c c}
    \hline
    Models & NYC-P & NYC-D & CPH-BS \\ [0.2ex] 
    \hline
    ConvLSTM & -352962 & -350803 & -112548 \\ [0.15ex]
    \textbf{RFN (Quantized)} & \textbf{-339745} & \textbf{-349627} & \textbf{-110999} \\[0.15ex]
    \hline
    \end{tabular}%
    \caption{Test log-likelihood for each task under the discrete support assumption. For the RFN, results are given after a quantization procedure mapping from a continuous 2d space to the $64 \times 64$ pixel space used to train the ConvLSTMs. Negative log-likelihood values are to be expected as these are evaluated under a discrete distribution, specifically a Categorical distribution.}
  \label{tab:results_discrete}%
\end{table}

\section{Conclusion}
\label{sec:conclusion}

This work addresses the problem of continuous spatio-temporal density modeling of urban mobility services by proposing a neural architecture leveraging the disentanglement of temporal and spatial variability. 
We propose the use of (i) recurrent latent variable models and (ii) conditional normalizing flows, as a general approach to represent temporal and spatial stochasticity, respectively.
Our experiments focus on real-world case studies and show how the proposed architecture is able to accurately model spatio-temporal distributions over complex urban topologies on a variety of scenarios.
Crucially, we show how the purposeful consideration of these two sources of variability allows the RFNs to exhibit a number of desirable properties, such as long-term stability and effective estimation of fine-grained distributions in longitude-latitude space.
From a practical perspective, we believe current mobility-on-demand services could highly benefit from such an additional geographical granularity in the estimation of user demand, potentially allowing for operational decisions to be taken in strong accordance with user needs.

\smallskip As a consideration for future research, this line of work could highly benefit in addressing two of what we believe to be the major limitations of the RFN.

First, complexity of probabilistic inference. 
Being defined as a latent variable model, the RFN faces the challenge of executing probabilistic inference within a class of models for which likelihood estimation cannot be computed exactly. 
In our work, we use techniques belonging to the field of (amortized) variational inference and demonstrate how these achieve satisfying results.
However, probabilistic inference is currently a very active area of research, known to be prone to common pitfalls and challenges e.g., posterior collapse \cite{LucasEtAl2021}, thus requiring extra care when compared to de-facto methods like traditional MLE.
Motivated by this, advances in the field of probabilistic inference could enable the widespread application of deep latent variable models, such as the RFN, within the demand forecasting literature.

Second, limitations of affine coupling layers. 
In our specific implementation of the RFN, the conditional output distribution is characterized by a sequence of affine coupling layers. 
At a high level, an affine coupling layer describes a transformation consisting of scale and translation operations, over a sub-set of the dimensions.
Intuitively, the scale and translation function are responsible for spreading (scale operation) and moving (translation operation) the support of the output distribution in space. 
Therefore, affine coupling layers might require a large number of transformations to represent distributions with disjoint support. 
In the case of urban mobility density estimation, being able to represent distributions with disjoint support represents the ability to model disjoint areas of user demand, thus avoiding the allocation of unnecessary probability mass in non-active areas.
In light of this, exploring different normalizing flow architectures could bring to relevant improvements in the accuracy of the resulting probability densities.

\smallskip From a transportation engineering perspective, we plan to incorporate the predictions of RFNs within downstream MoD supply optimization routines, with the goal of enabling better resource allocation (e.g. vehicle rebalancing, inventory management, etc.) and increased demand satisfaction.

\bibliographystyle{plain}
\bibliography{paper2_bib}

\newpage
\appendix 
\section{Evidence Lower BOund Derivation}
We herby report the derivation to obtain the evidence lower bound used to train the proposed RFN in Eq.\ref{eq:elbo}:
\begin{align}
    \log p_{\theta}(\mathbf{x}_{1:T}) & = \log \int p_{\theta}(\mathbf{x}_{1:T}, \mathbf{z}_{1:T}, \mathbf{h}_{1:T}) d\mathbf{z} \, d\mathbf{h} \nonumber\\
    & = \log \int \frac{q_{\phi}(\mathbf{z}_{1:T} | \mathbf{x}_{1:T})}{q_{\phi}(\mathbf{z}_{1:T} | \mathbf{x}_{1:T})} p_{\theta}(\mathbf{x}_{1:T}, \mathbf{z}_{1:T}, \mathbf{h}_{1:T}) d\mathbf{z} \, d\mathbf{h} \nonumber\\
    & = \log \mathbb{E}_{q_{\phi}(\mathbf{z}_{1:T} | \mathbf{x}_{1:T})} \left[\prod_{t=1}^{T} \frac{p_{\theta}(\mathbf{x}_{t} | \mathbf{z}_{t}, \mathbf{h}_{t}) p_{\theta}(\mathbf{z}_{t} | \mathbf{z}_{t-1}, \mathbf{h}_{t}) p_{\theta}(\mathbf{h}_{t} | \mathbf{h}_{t-1}, \mathbf{u}_{t})}{q_{\phi}(\mathbf{z}_{t} | \mathbf{z}_{t-1}, \mathbf{h}_{t}, \mathbf{x}_{t})} \right] \nonumber\\
    & \geq \mathbb{E}_{q_{\phi}(\mathbf{z}_{1:T} | \mathbf{x}_{1:T})} \left[\sum_{t=1}^{T} \log p_{\theta}(\mathbf{x}_{t} | \mathbf{z}_{t}, \mathbf{h}_{t}) + \log p_{\theta}(\mathbf{h}_{t} | \mathbf{h}_{t-1}, \mathbf{u}_{t}) + \log \left( \frac{p_{\theta}(\mathbf{z}_{t} | \mathbf{z}_{t-1}, \mathbf{h}_{t})}{q_{\phi}(\mathbf{z}_{t} | \mathbf{z}_{t-1}, \mathbf{h}_{t}, \mathbf{x}_{t})} \right) \right] \nonumber\\
    & = \mathbb{E}_{q_{\phi}(\mathbf{z}_{1:T} | \mathbf{x}_{1:T})} \left[\sum_{t=1}^{T} \log p_{\theta}(\mathbf{x}_{t} | \mathbf{z}_{t}, \mathbf{h}_{t}) + \log p_{\theta}(\mathbf{h}_{t} | \mathbf{h}_{t-1}, \mathbf{u}_{t}) \right] \nonumber\\
    & - \sum_{t=1}^{T} \mathbb{KL} \left(q_{\phi}(\mathbf{z}_{t} | \mathbf{z}_{t-1}, \mathbf{h}_{t}, \mathbf{x}_{t}) || p_{\theta}(\mathbf{z}_{t} | \mathbf{z}_{t-1}, \mathbf{h}_{t}) \right) = \mathcal{L}(\theta, \phi) \nonumber
\end{align}

\section{Experimental details}
For every model considered under the continuous support assumption, we select a single layer of 128 LSTM cells.
The feature extractor $\varphi^{\text{extr}}_{\tau}$ in Eq. \eqref{eq:rfn_t_1} has three layers of 128 hidden units using rectified linear activations \cite{NairEtAl2010}.
For the VRNN, SRNN and RFN we also define a 128-dimensional latent state $\bz_{1:T}$. Both the transition function $t_{\theta_z}$ from Eq. \eqref{eq:rfn_t_2} and the inference network $\varphi^{\text{enc}}_{\tau}$ in Eq. \eqref{eq:rfn_inf} use a single layer of 128 hidden units. 
For the mixture-based models, the MDN emission is further defined by two layers of 64 hidden units where we use a softplus activation to ensure the positivity of the variance vector in the MDN-Diag case and a Cholesky decomposition of the full covariance matrix in MDN-Full.
Based on a random search, we use 50 and 30 mixtures for MDN-Diag and MDN-Full respectively.
The emission function in the RFN is defined as in Eq. \eqref{eq:rfn_prior} and Eq. \eqref{eq:rfn_flow}, where $f_{\psi}$, $s_{\psi}$ and $t_{\psi}$ are neural networks with two layers of 128 hidden units. The conditional flow is further defined as an alternation of 35 layers of the triplet [Affine coupling layer, Batch Normalization \cite{IoffeSzegedy2015}, Permutation], where the permutation ensures that all dimensions are processed by the affine coupling layers and where the batch normalization ensures better propagation of the training signal, as shown in \cite{DinhEtAl2017}. In our experiments we define $\bu_t = \bx_{t-1}$, although $\bu_t$ could potentially be used to introduce relevant information for the problem at hand (e.g. weather or special event data in the case of spatio-temporal transportation demand estimation).

\smallskip \noindent On the other hand, under the discrete support assumption, we train a 5-layer ConvLSTM network with 4 layers containing $40$ hidden states and $3 \times 3$ kernels (in alternation with $4$ batch normalization layers) using zero-padding to ensure preservation of tensor dimensions, a $3$D Convolution layer with kernel $3 \times 3 \times 3$ and softmax activation function to describe a normalized density over the next frame (i.e. time-step) in the sequence.

\smallskip \noindent All models assuming continuous output distribution were implemented using PyTorch \cite{PaszkeGrossEtAl2019} and the universal probabilistic programming language Pyro \cite{BinghamEtAl2018}, while the ConvLSTMs where implemented using Tensorflow \cite{Abadi2015}. To reduce computational cost, we use a single sample to approximate the intractable expectations in the ELBO.

\end{document}